\documentclass{article}
\usepackage{ijcai19}
\usepackage{makecell}
\usepackage{times}
\usepackage{soul}
\usepackage{url}
\usepackage[hidelinks]{hyperref}
\usepackage[utf8]{inputenc}
\usepackage[small]{caption}
\usepackage{graphicx}
\usepackage{subcaption}
\usepackage{amsmath}
\usepackage{booktabs}
\usepackage{multirow}
\usepackage{todonotes}

\title{Approximating Optimisation Solutions for Travelling Officer Problem with Customised Deep Learning Network}

\author{
Wei Shao$^1$\footnote{Contact Author}\and
Flora D. Salim$^2$\and
Jeffrey Chan$^{2,3}$\And
Sean Morrison$^4$\\
\affiliations
$^1$RMIT University\\
\emails
\{first, second\}@example.com,
third@other.example.com,
fourth@example.com
}

\begin{document}

\maketitle

\begin{abstract}
Deep learning has been extended to a number of new domains with critical success, though some traditional orienteering problems such as the Travelling Salesman Problem (TSP) and its variants are not commonly solved using such techniques. Deep neural networks (DNNs) are a potentially promising and under-explored solution to solve these problems due to their powerful function approximation abilities, and their fast feed-forward computation. In this paper, we outline a method for converting an orienteering problem into a classification problem, and design a customised multi-layer deep learning network to approximate traditional optimisation solutions to this problem. We test the performance of the network on a real-world parking violation dataset, and conduct a generic study that empirically shows the critical architectural components that affect network performance for this problem.
\end{abstract}

\section{Introduction}

\begin{figure}[ht]
	\centering
	\includegraphics[width=0.8\linewidth]{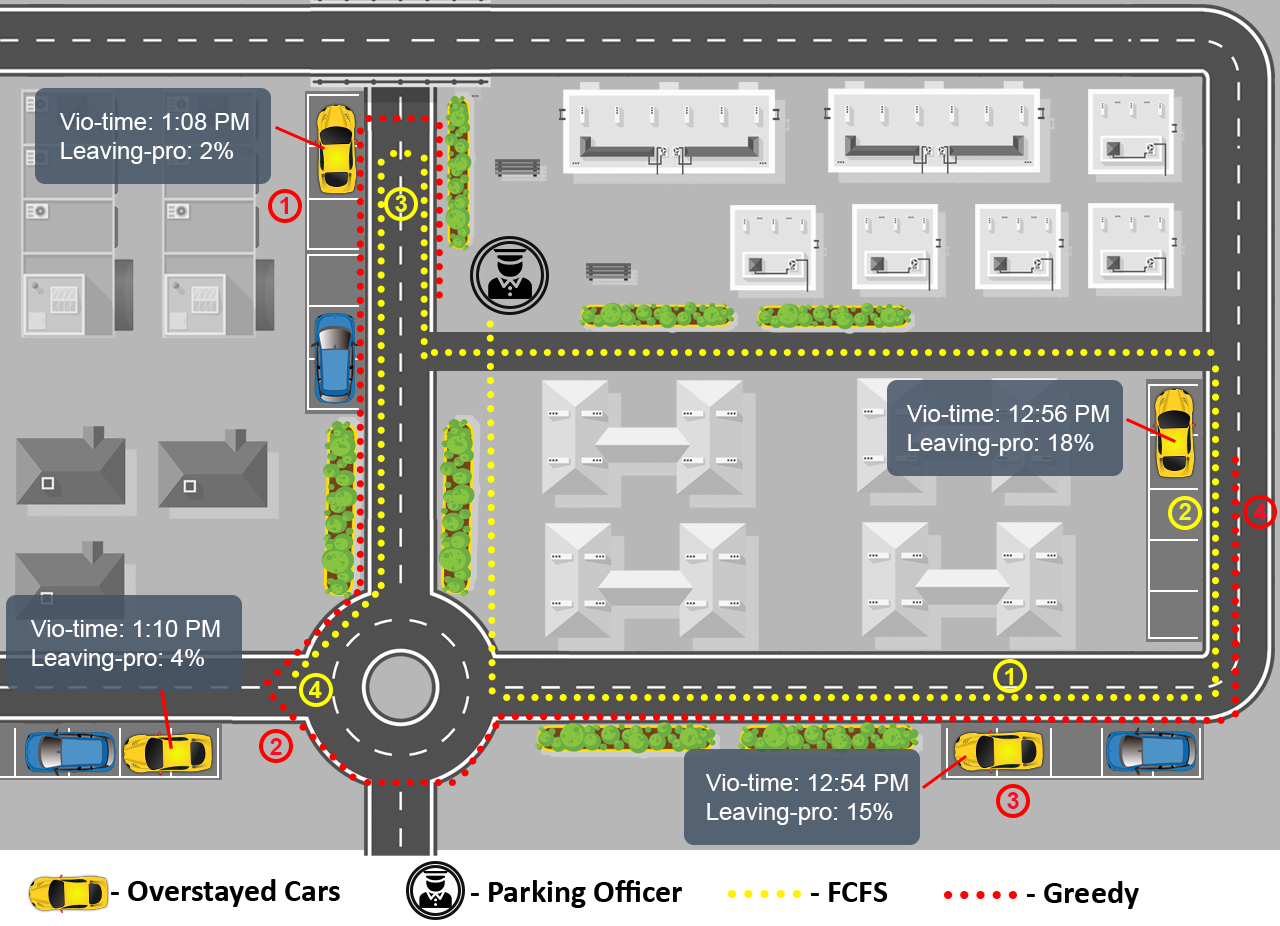}
\caption{The Travelling Officer Problem -- a parking officer faces four parking violation in different directions. The parking officer knows the time-in-violation for each car, and estimates the probability of each car leaving. The parking officer needs to choose a path that catches as many of these violators as possible. Yellow line in the map shows the First-Come-First-Serve (FCFS) solution. The red line illustrates the probability-based greedy solution. } \label{fig_TOP}
\end{figure}

The travelling officer problem (TOP) is a variant of travelling salesman problem (TSP) \cite{shao2017travelling} but provides a way to use contextual and historical data. Nowadays, parking violation has become a prominent challenge for administration in most big cities. Parking officers need to stick infringement notices on violating cars before they leave the parking zone, but this can be challenging for many reasons. Firstly, the majority of infringing vehicles leave within a short period. Secondly, many violation events occur at the same time in a large area. As shown in Figure \ref{fig_TOP}, parking officers must balance the travelling time between the officer's location relative to the infringing vehicles, and the probability that these cars will leave. 

Shao et al. \cite{shao2017travelling} previously defined this problem, and two heuristic solutions were demonstrated for generating paths using spatio-temporal data collected from on-ground sensors in parking spaces. The path generated by these optimisation methods (E.g. Red line in Figure \ref{fig_TOP} ) were shown to be better than the First-Come-First-Serve solutions (E.g. Yellow line in Figure \ref{fig_TOP}), and achieved a higher return on parking fines when evaluated on the real-world parking dataset \cite{shao2017travelling}. Despite this, the efficiency of these optimisation solutions cannot satisfy the requirements of real-time application. Given the short-term nature of parking violations, this makes application of more traditional methods difficult.

Deep learning models can achieve real-time performance since their training and inference are broken up into two distinct sessions. Training can be done offline, and inference runs in constant time, making neural networks ideal for problems like the TOP where fast evaluation is desired. However, a significant portion of current classification approaches tend to solve supervised learning problems for which the solution is known and provided as a training dataset. In addition, the TOP is a typical optimisation problem, and deep learning models are not typically used to solve optimisation problems directly. In order to leverage deep learning models, the TOP problem needs to be transformed into a supervised classification problem in order to obtain good potential paths as labels.

To overcome the above challenges, we use solutions generated by optimisation methods as labels for training. We propose a spatio-temporal data segmentation approach to transform the optimisation problem into a classification problem, and design a deep feed-forward neural network to approximate optimisation solutions. Training deep neural networks to replace optimisation has many advantages in this case: the computationally expensive optimisation problem can be solved as part of the training session, and once this is done, the test session can roll out trajectories that approximate those of the original optimiser with simple feed-forward computation. Moreover, neural networks scale very well, allowing such a technique to take advantage of a huge amount of additional contextual and temporal information and explore the unclear structure of this data. 

We choose deep neural networks (DNN) rather than traditional machine learning classifiers such as Random Forest \cite{breiman2001random} and SVM \cite{suykens1999least} because DNN can be customised to our purpose. To approximate specific optimisation solutions, we can customise different DNNs which can learn each operation of the specific optimisation method.

There are challenges in both the transformation and learning tasks. Traditional orienteering problems only focus on the spatial domain, usually in the form of a static 2D graph. The TOP also considers the temporal domain, which consists of many temporal views of this same spatial graph. We need a method that works with both spatial and temporal information. Secondly, classification methods need the input features and the corresponding labels; the TOP does not provide any existing solutions or features. It is non-trivial to integrate the optimisation solution with classification to solve this problem. Finally, it is unclear how to effectively use deep neural networks to approximate the optimisation solution in orienteering problems. As a result, the design of an appropriate deep learning architecture for this type of problem is needed. 

This paper explores the use of deep learning techniques for the solution of the TOP, and makes the following contributions:

\begin{itemize}
\item We propose a generic framework to solve travelling officer problem incorporating optimisation approaches and deep neural networks.
\item We propose a novel segmentation method to transform a spatio-temporal graph into a sequence of features. 
\item We are the first group to customise a neural network to approximate the greedy algorithm.
\item We validate our claim that the TOP can be solved using a combination of both optimisation solutions and neural network classifiers through extensive experiments with a large real-world dataset, including a comparison with traditional machine learning methods. 
\end{itemize}

The paper is organised as follows: Section \ref{relatedwork} discusses the related work; some preliminary studies are shown in Section \ref{chapter:preliminarystudies}; Section \ref{chapter:method} presents the problems and corresponding methodologies in both data representation and deep learning architecture; Section \ref{chapter:ExperimentAndResults} shows the experiments and comparison studies; Section \ref{chapter:DiscussionAndFutureWork} discusses the limitations of this method, and future work; Section \ref{chapter:conclusion} concludes the paper.

\section{Related Work}
\label{relatedwork}
Related work in this area falls into two categories: 1) Neural network solutions for the TSP, and 2) general work in the intersection of deep learning and optimisation.

There is an extensive body of research in applying neural networks to TSP variants going back to \cite{hopfield1985neural}, though the networks used in these studies typically fall into the category of Hopfield networks, and self-organising feature map networks (see \cite{la2012comparison,abdel2010traveling,potvin1992traveling}). Hopfield networks are fully recurrent, and memorise training examples by minimising an energy cost function. However, many recent advances in deep learning have been made with feed-forward neural networks, on the back of better optimisation algorithms and the ability to train on large datasets. Though Hopfield networks have been used for classification tasks, their performance is not as good as modern deep learning techniques. On combinatorial tasks, the number of neurons required by a Hopfield network scales with $n^2$ (where $n$ is the number of nodes in the graph) which can be problematic for larger graphs. Our study is different from these previous works in two ways: firstly, we are approximating a solution to a more difficult, time-dependent version of the TSP (the TOP). Our problem is focused on maximising a temporally-dependent reward, rather than navigating a geographically-fixed set of nodes. Using different temporal views of the data to generate training samples can dramatically increase the size of the training set (this is inspired by \cite{liu2016urban}). Secondly, rather than using Hopfield networks, we propose re-framing the problem as a supervised learning task for classification. Under this framework, we use an optimisation algorithm to generate the training set for a classifier, which is then trained to generate a trajectory through a given graph.

Additionally, there numerous recent works in effectively combining optimisation with deep learning. Fischetti and Jo modelled deep neural networks as a 0-1 mixed integer linear program \cite{fischetti2017deep}. Galassi et al. used a deep neural net to learn the structure of a combinatorial problem, and mentioned that such research is still at an early stage \cite{galassi2018model}. Our work makes a small contribution to this area.

\section{Background}\label{chapter:preliminarystudies}
\subsection{Travelling Officer Problem}
The Travelling Officer Problem describes the problem of a parking officer traversing a fully connected graph to maximise a cumulative reward (in this case, parking violations). There is a time cost $\mathrm{C}(\mathit{u}\boldsymbol{,\ }\mathit{w})$ associated with travelling from node $\mathit{u} \in \boldsymbol{v}$ to $\mathit{w} \in \boldsymbol{v}$ (or parking lot $\mathit{u}$ to $\mathit{w}$, where $\boldsymbol{v}$ denotes all nodes in the graph or all parking lots in the area) that is dependent on the officer's walking speed (we do not assume that the officer stops at intermediary nodes).

The officer must choose between chasing for the potential reward for catching a parking violation at a given node by considering the probability that the violation may no longer exist by the time he/she arrives, and the opportunity of cost saving from not travelling to other nodes containing parking violations. The solution of the TOP aims to find a path $\mathbf{S}$ that maximises the number of valid nodes with time limits (e.g. working hours), and the time-varying state of each node. The valid nodes denote the car at parking lot $x_i$ is in a state of violation.


Let $\mathit{T}$ be the total travelling budget, and $\mathit{f}_{j,t} \in \{0,\ 1\}$ denotes whether there is an infringement at node $\mathit{j}$ at time $\mathit{t}$. We denote a solution $\mathbf{S} = \{(\mathit{x_1}, \mathit{t_1}), (\mathit{x_2}, \mathit{t_2}),..., (\mathit{x_{|\mathbf{S}|}, \mathit{t_{|\mathbf{S|}}})}\}$ as the path travelling over nodes, where $\mathit{x_i} \in \boldsymbol{v}$ and denote the $i_{th}$ node in the path, and $\mathit{t_i} \in Time$ denote the time when officer arrives at the node $\mathit{x_i}$ (note Time is whatever time unit/division you are using). Because in the TOP, $\mathit{t_i}$ is deterministic from the path of nodes visited, we can infer $t_i$ from just the visited nodes, and we will simplify our path $\mathbf{s}$ to $x_1, x_2, \dots, x_{|S|}$. Let $\mathit{R}$ denote the infringement fine amount (assuming each infringement cost the same). In this paper, we assume the $\mathit{R}$ is a constant value.

Then the TOP problem is to find a path S that maximises the total return, satisfying the total travelling time budget. A formal definition of this problem is as follows:

\begin{eqnarray}
\nonumber  & \underset{\mathbf{S}}{\arg\max \ } \ \sum\limits _{\mathit{x}_{i} \in \mathbf{S}} f_{i,t} \cdot \mathit{R}  \\
\nonumber \text{s.t.} & \\
&\sum\nolimits ^{|\mathbf{S}|-1}_{i=1} \mathrm{C}( x_{i} ,\ x_{i+1}) \ \leq \mathit{T} \\
& \mathit{t}_{k} =\sum\nolimits ^{k}_{i=2}\mathrm{C}( x_{i} ,\ \mathit{x}_{i+1}), \text{for } 1 < k < |\mathbf{S}| \\
& t_{ 1 }= 1   
\end{eqnarray}

\subsection{Heuristic Optimisation}
Previously, Shao et al. \cite{shao2017travelling} discussed two heuristic optimisation methods (greedy and ACO) to solve the Travelling Officer Problem; it was shown that both algorithms performed well at the task of collecting parking violation fines. In order to take advantage of temporal information in the TOP, the authors proposed a dynamic temporal probability model and integrated it with traditional optimisation methods. The proposed greedy algorithm can be formalised as a single function as follows:
\begin{equation} \label{eq:5}
 \underset{\mathit{i}}{\arg\max}\exp\left( -\frac{\tau _{i} +\frac{\mathrm{d}_{i}}{V}}{\alpha }\right),
\end{equation}
where $\tau_i$ denotes the overstayed time of cars at node $\mathit{x_i}$, and $\mathrm{d_i}$ denotes the route distance between node $\mathit{x_i}$ and the current position of the parking office. $V$ is a constant to denote the speed of the parking officer, and $\alpha$ is a parameter which is set by historical data analysis. The proposed greedy algorithm seek for $i_{th}$ nodes calculated by Eq. \ref{eq:5} as the next position for parking officer.

In the ACO algorithm, the ants decide the next node by the pheromone distribution left by previous ants. The probability of these choices was modelled as being proportional to ${[\phi ({x_{uw}})]^\alpha } \bullet {[\eta ({x_{uw}})]^\beta }$, where $\eta$ is the probability of a node being invalid by the time the officer arrives. The greedy algorithm used a similar dynamic probability model to estimate the most promising node, and would greedily select the best one. The details of both algorithms are shown in \cite{shao2017travelling}.

\begin{figure*}[!ht]
	\centering
	\includegraphics[width=0.8\textwidth]{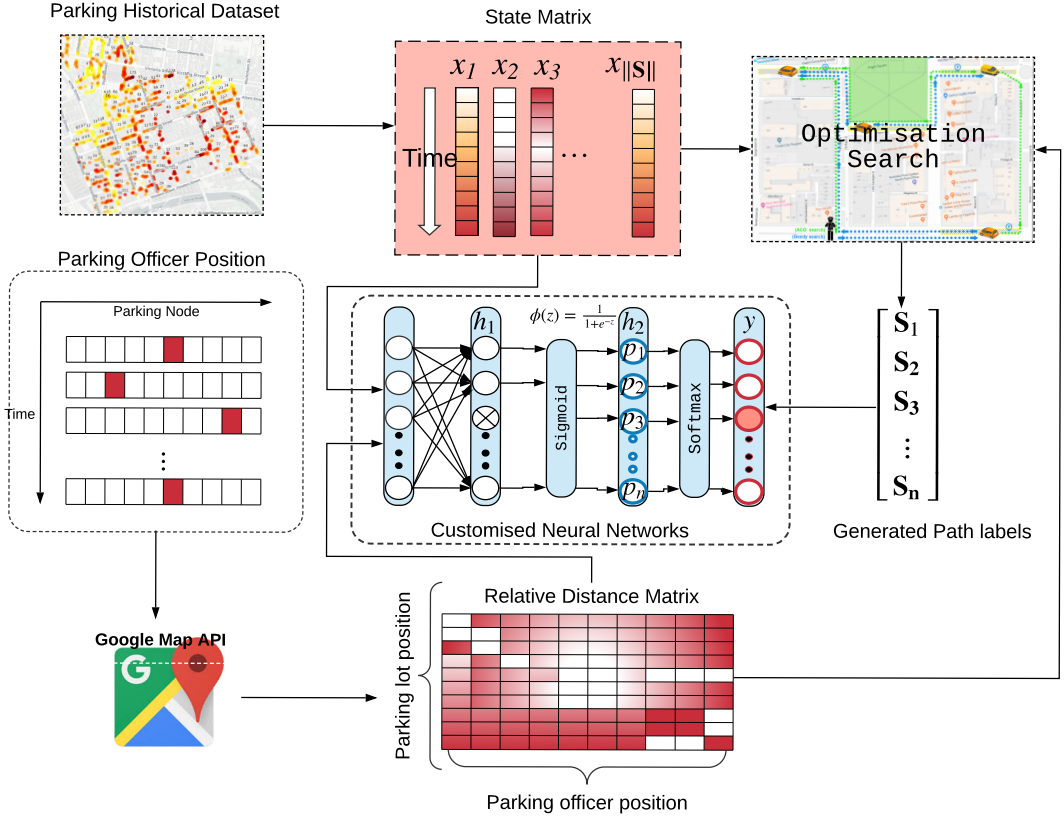}
	\caption{Flowchart of the proposed model. The state matrix is extracted from parking event data. The officer position matrix is randomly generated, and a relative distance matrix which measures the distance between the parking officer and other parking lots is calculated using the Google Maps API. From there, the state matrix and relative distance matrix are fed into the optimiser to generate labels for the classifier. These labels are then used for training, and the classifier is used to roll out trajectories on the test set.} \label{fig:overview}
\end{figure*}

\section{Methodology}
\label{chapter:method}
In this section, we propose a framework to transform the TOP -- a spatio-temporal orienteering problem -- into a classification problem that can be solved by a customised deep neural networks incorporating optimisation solutions. Figure \ref{fig:overview} illustrates the overview of our framework. It primarily consists of three parts: spatio-temporal feature extraction, optimisation-based search, and a customised neural network. Spatio-temporal feature extraction aims to extract features and build a training dataset from the public parking violation historical dataset. Classification labels are generated by using optimisation to find a path from any given map. The DNN then uses the state and relative distance as the input features, and uses the optimisation solution $\mathbf{S}$ as the ground truth label to learn the optimisation algorithm. We outline the details of each component in the following subsections. 

\subsection{Spatio-temporal Feature Extraction Method} \label{sec:st_dfem}
A significant difference between traditional sub-path planning problems and the TOP is temporal dependence. Traditional sub-path problems generally consist of spatial data such as the location of the vertices, or the edges of the graph. For such static graphs, information is limited; however, time is continuous, meaning that even finer-grained slices can be taken to generate further data points when the temporal dimension is included. Each time frame becomes a static 2-D graph at timestamp $t_i$, corresponding to a data sample $\mathit{\chi }_i$.

\begin{figure}[ht]
	\centering
	\includegraphics[width=0.8\linewidth]{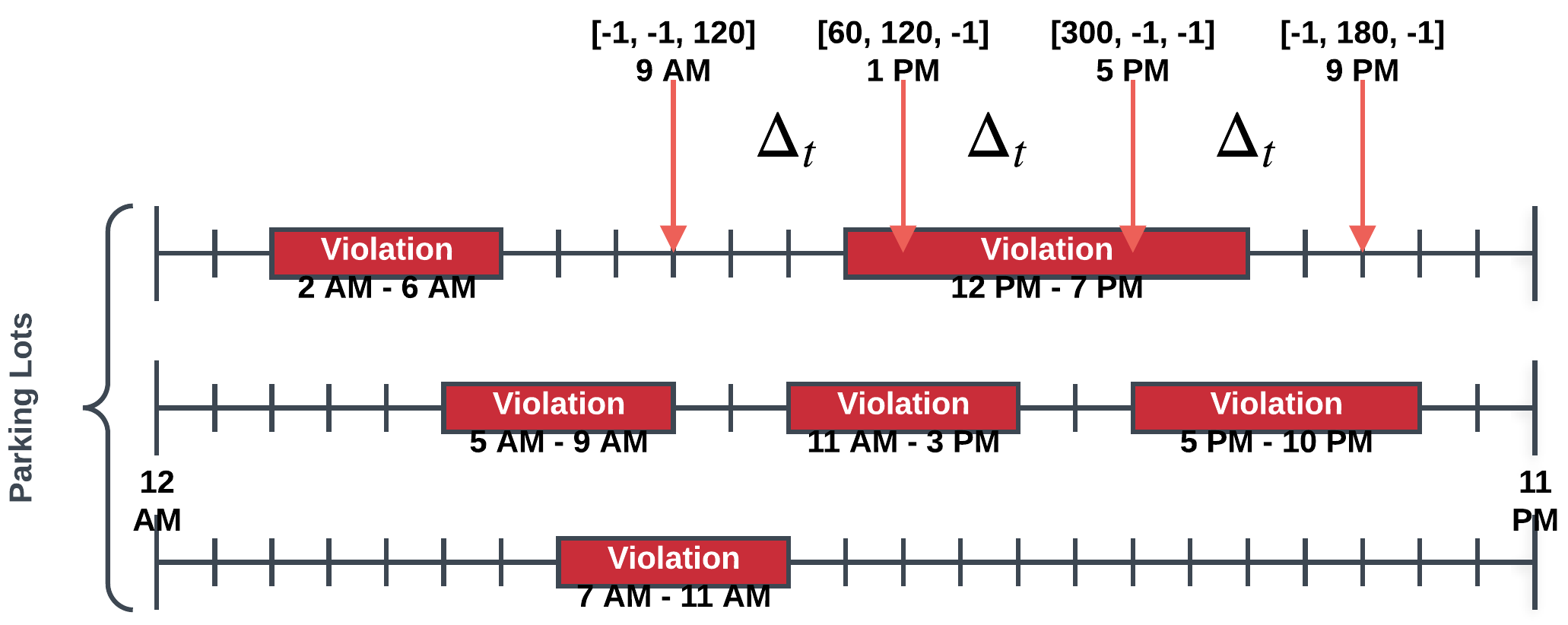}
	\caption{Spatio-temporal feature extraction method: The bottom part represents the parking violation events in temporal domain. The red bar indicates the violation interval. The red arrows slices the state information of each node with a fix-length time step. Inf denotes that there is no violation in the specific parking lot. } \label{fig:feature_extraction}
\end{figure}

As outlined in Figure \ref{fig:feature_extraction}, we extract a vector $\chi_i$ = $\{\chi_{i,x_1}, \chi_{i,x_2},\dots,\chi_{i,x_n}\}$, where $n$ is the number of parking lots (the number of nodes in the graph) from parking violation events at time $t_i$ and slicing time frame every $\Delta_t$.

For example, at node $x_j$, there are $p$ violation events $\{(\tau^a_{x_j, 1}, \tau^e_{x_j, 1}), (\tau^a_{x_j, 2}, \tau^e_{x_j, 2}), \dots, (\tau^a_{x_j, p}, \tau^e_{x_j, p})\}$, where $\tau^a_{x_j, k}$ is the start time of the $k_{th}$ violation interval at node $x_j$, and $\tau^e_{x_j, k}$ is the end time of the $k_{th}$ violation interval at node $x_j$. For any time $t_i$, there only two cases: 1) $t_i$ is located between one of violation events interval. That is, $\tau^a_{x_j, k} < t_i < \tau^e_{x_j, k}$ and $1 \leq k \leq p$, or 2) time $t_i$ is not located in any violation events. In this first case, we set the value $\chi_{i, x_j} = t_i - \tau^a_{x_j, k}$. In the second case, We set the value as $-1$ if there is no violation for parking lot $x_j$ at time $t_i$. The number of time lots is denoted as $m = \frac { T }{ \Delta_t } $, where $T$ is the maximum time constraint and $\Delta_t$ denotes the time step we choose for slicing the temporal domain. 

Except for temporal features, we also extract spatial features. The first component in the second row of Figure \ref{fig:overview} shows a officer position matrix, where position filled with red colour in each row denotes the current location of the parking officer at time $t_i$. For simplicity, the officer will not change the path between two nodes. Therefore, the possible position of officers are the same as the parking lots positions. The relative distance matrix shown in the bottom row in the Figure \ref{fig:overview} store the distance between current parking officer location and other nodes $x_j \in \boldsymbol{v}$. Therefore, for any time $t_i$, we can get a relative distance vector $\boldsymbol{d} = \{\mathit{d_1}, \mathit{d_1}, \dots, \mathit{d_n}\}$, where $\mathit{d_j}$ measure the route distance between the parking officer and node $x_j$. 

\subsection{Optimisation}
The TOP is an NP-hard problem that can be solved by traditional optimisation methods under an objective function and constraints. As mentioned previously, optimisation solution $\mathbf{S}$ is series of nodes at time $t_i$. However, a complete path should be broken into multiple decisions to transform an optimisation problem to a classification problem. That is, at any time $t_i$, we need to run the optimisation algorithm and only select the first node $x_1$ in the path $\mathbf{S}$. As a result, we can get a series of labels $\boldsymbol{l} = \{x^{t_1}_1, x^{t_2}_1,\dots, x^{t_m}_1\}$, where $x^{t_i}_1$ is the first node of path $\mathbf{S}$ which is given by optimisation method at time $t_i$. 

The classification problem aims to learn a categorical likelihood over a set of classes from a training set. In this case, if we regard each node $x_i$ as a class label and use state matrix and relative distance matrix as the features for the training, the classifier should learn to choose the next node.

\subsection{Customised Neural Networks}
\begin{figure}
	\centering
	\includegraphics[width=1.0\linewidth]{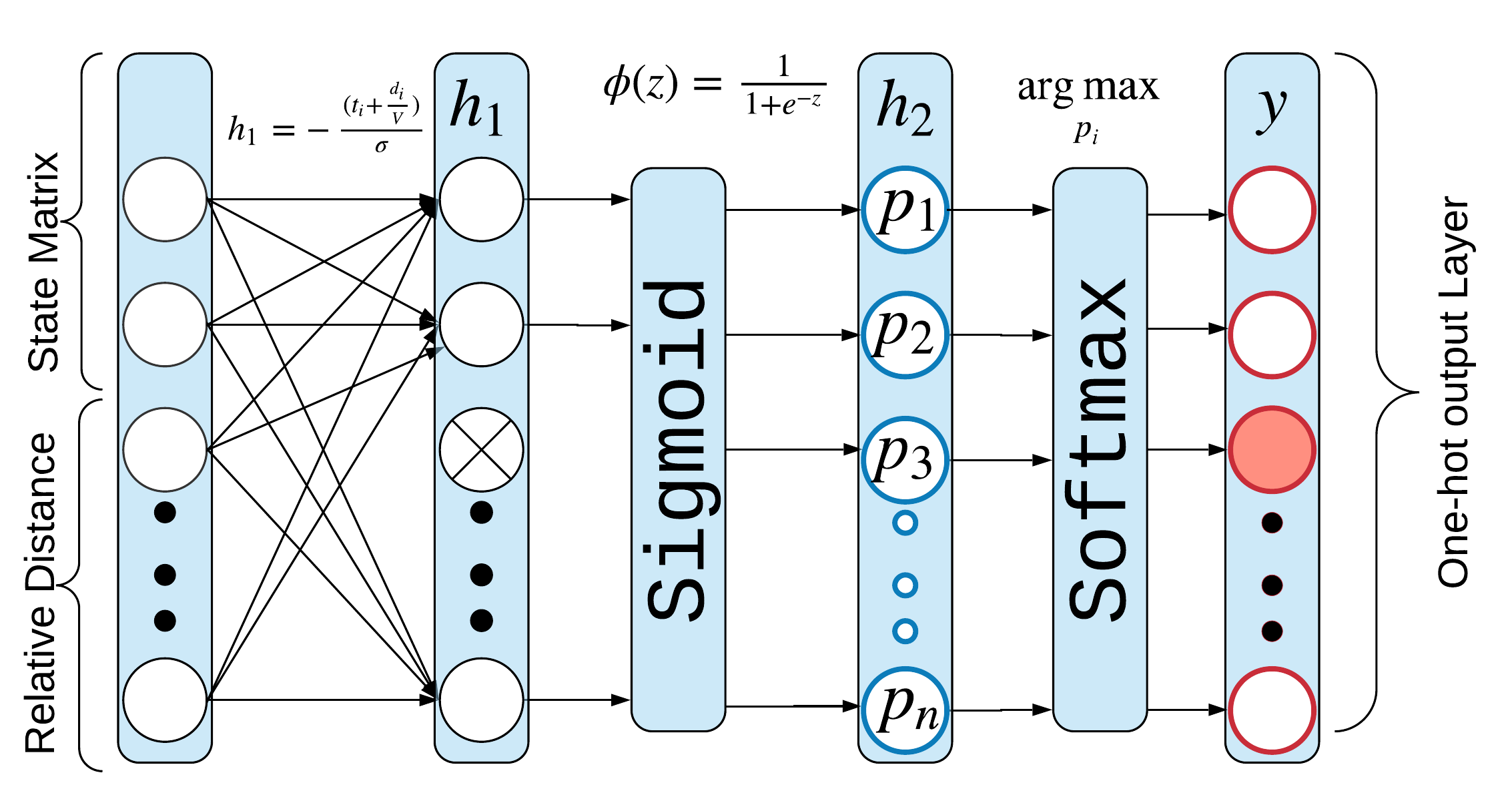}
	\caption{Architecture of greedy-like DNN. Input layer consists of relative distance vector and state vector for each parking node. Output layer is a one-hot vector denotes the next travelling node.} \label{fig:DNN}
\end{figure}

In order to achieve similar performance with optimisation methods, it is not enough to use an existing general neural network because the general neural networks are sensitive to the hyper-parameters and architecture. Therefore, we propose a customised neural network which is designed with each operation of the greedy algorithm in TOP. Figure \ref{fig:DNN} shows the completed architecture of neural networks designed for replacing the greedy algorithm in TOP. The first layer is the input layer concatenated by two vectors: a relative distance vector $\boldsymbol{d}$, and the state vector $\chi_{i}$ which is calculated in Section \ref{sec:st_dfem}. The first hidden layer $h_1$ aims to learn a linear combination of $\boldsymbol{d}$ and $\chi_i$. If we pay attention to Eq. \ref{eq:5}, we can find that this layer can approximate the function $\mathrm{h_1}(\mathrm{d}_{i},\tau _{i}) =  -\frac{\tau _{i} +\frac{\mathrm{d}_{i}}{V}}{\alpha }$. Universal approximation theorem \cite{csaji2001approximation} states that a feed-forward network with a single hidden layer containing a finite number of neurons can approximate continuous functions on compact subsets of $\mathbf{R}^n$. This is a continuous function on a real value dataset. Therefore, we design a hidden layer $h_1$ to approximate this function. Then we use a sigmoid function which is also a non-linear activation function to approximate exponential function. This is also a non-linear function \cite{ito1991approximation}\cite{ferrari2005smooth}. Layer $h_2$ now consists probabilities $p_i$ to denote the capture chance by parking officer for node $x_i$. Since $\arg\max$ is not a continuous function, we cannot use the hidden layer. Fortunately, softmax function is a perfect function to choose the max value from $\boldsymbol{p} = \{p_1, p_2,\dots, p_n\}$ \cite{tokic2011value}. Therefore, the output of the network becomes the next node that the officer should travel to.  

For other components of the neural network, we used Adam \cite{kingma2014adam} for optimisation, and early stopping for regularisation. We used dropout at each layer to prevent overfitting \cite{JMLR:v15:srivastava14a}. 

\section{Experiments and Results}\label{chapter:ExperimentAndResults}
In this section, we evaluate the performance of our proposed model, and compare it with traditional classification methods and optimisation-only solutions on a real-world dataset. The rules and assumptions are outlined in \cite{shao2017travelling}.


\subsection{Dataset}
We tested the proposed model on the Melbourne parking event dataset, published by the Melbourne City Council, and used previously in \cite{shao2016clustering}, \cite{shao2017travelling}.  A detailed description of this dataset is included in \cite{shao2017travelling}. We took time slices throughout the week for training, and tested the performance of the classifier using different time slices to ensure that the test and train sets were drawn from the same distribution. For reward evaluation, we randomly chose a week's worth of data from a year-long data set. For other experiments, we randomly selected a single day. For the rewards study, we chose $50$ nodes from all vertices in the graph, and extracted sampling data at $10$ second intervals.

\subsection{Evaluation Metric}
We use two criteria to evaluate the performance of our proposed method: rewards, and classification accuracy. The definition of rewards is given in \cite{shao2017travelling}. It denotes how many cars in violation can be caught by parking officers. Since we use the optimisation solution as the ground truth, we also use the classification accuracy to measure the degree to which the classifier learns the optimisation algorithm. 

\subsection{Experimental settings}
We applied both ACO and greedy algorithm which is used in Shao et al. \cite{shao2017travelling} to the dataset and our customised DNN, Support vector machine (SVM) \cite{smola2004tutorial} and Random Forest (RF) \cite{breiman2001random} to learn the optimisation solutions.



\subsection{Classification Model Comparison}
In the first experiment, we evaluate all classifiers over a week by rewards and categorical accuracy. There are two sub-set of experiments. First experiment compares the greedy algorithms and classifiers that learn from the greedy algorithm. Second experiments evaluates the ACO and classifiers that learn from ACO.

\begin{figure}[ht]
\centering
	\includegraphics[width=0.8\columnwidth]{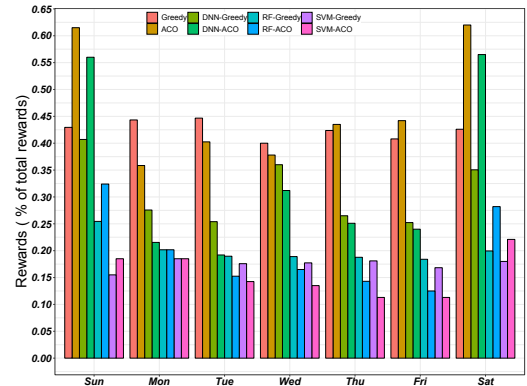}
	\caption{Weekly rewards obtained through different algorithms.} \label{fig_rewards}
\end{figure}

Figure \ref{fig_rewards} shows the weekly rewards obtained by optimisation solutions -- greedy and ACO. It also shows achieved rewards from different classification methods learned from the optimisation solutions. customised DNN outperformed the other techniques. Interestingly, we found that classification was more accurate on weekends compared to weekdays because weekends average a lower number of violations than weekdays, largely due to less stringent parking rules. Overall, The DNN achieves similar performance to the greedy on this problem as expected.


\subsection{Evaluation of Model Components}
We also studied the effect of varying the parameter settings of the problem, as measured by classification accuracy and rewards achieved:
\begin{itemize}
	\item Number of nodes: the number of nodes indicates the depth of the search space in orienteering problems. In this case, the number of nodes is also associated with the number of rewards, and the size of the training set;
	\item Minimum time step: we extract training samples from the dataset with a time step. For each time step $t_i$, we extract a temporal image from the dataset and add it to our training set.    
\end{itemize}

\begin{figure}
	\centering
	\begin{subfigure}[b]{0.49\columnwidth}\label{fig:rewards_nodes}
		\includegraphics[width=\columnwidth]{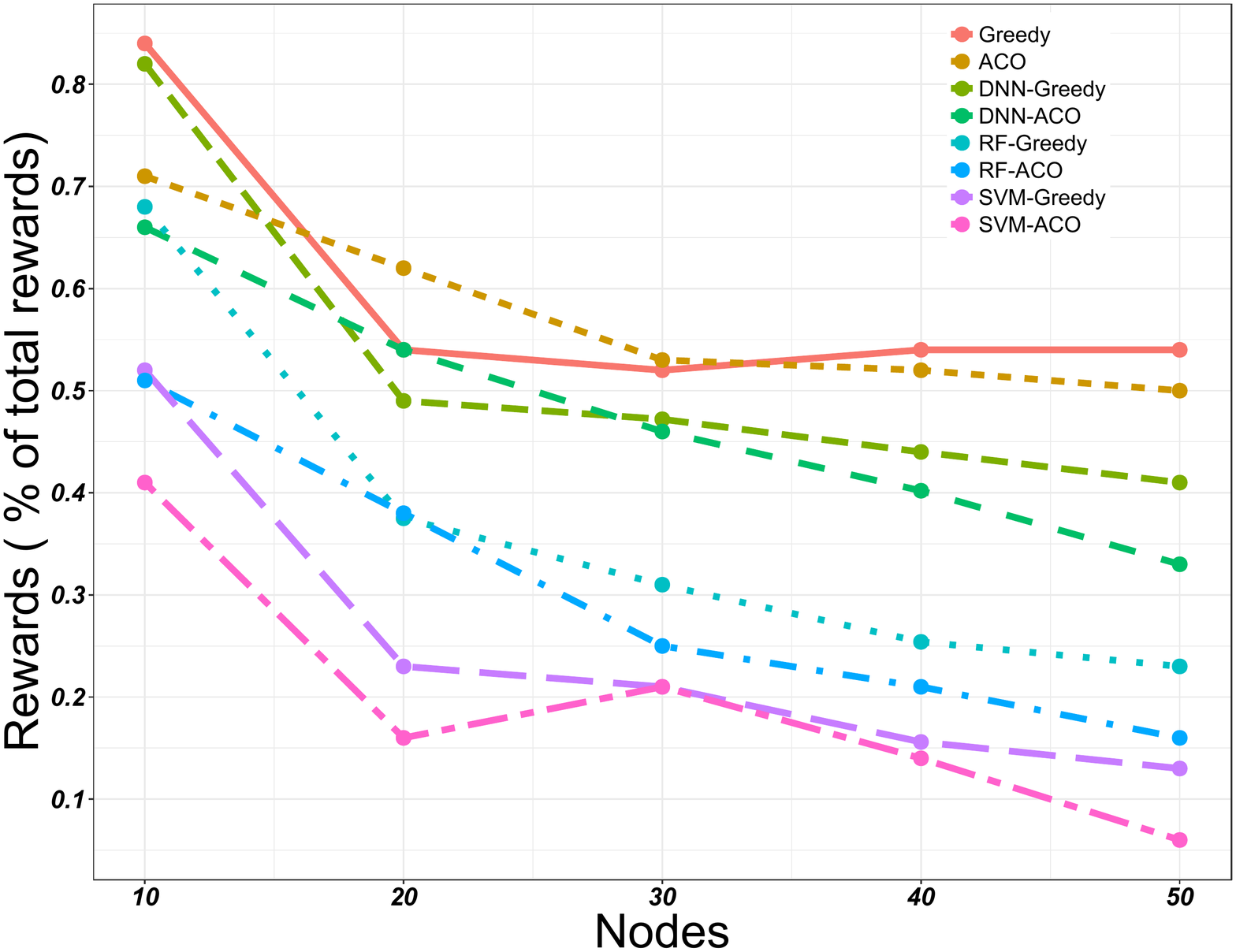}
	\end{subfigure}%
	\begin{subfigure}[b]{0.49\columnwidth}\label{fig:acc_nodes}
		\includegraphics[width=\columnwidth]{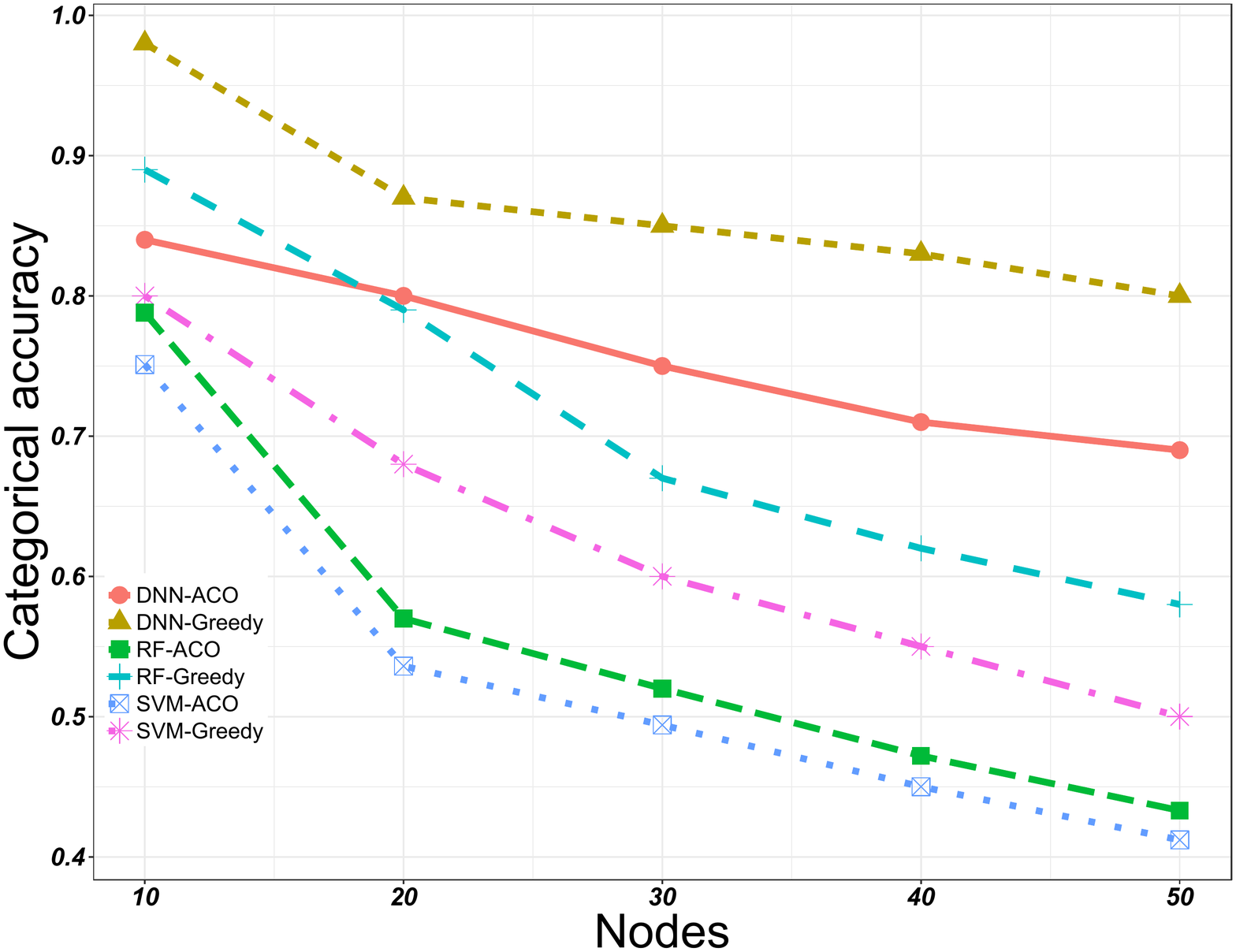}
	\end{subfigure}%
	\caption{The effect of the number of nodes $n$ on the total reward obtained, and the categorical accuracy as seen on the test set.}
	\label{fig:nodes}
\end{figure}

\begin{figure}
	\centering
	\begin{subfigure}[b]{0.49\columnwidth}
		\includegraphics[width=\columnwidth]{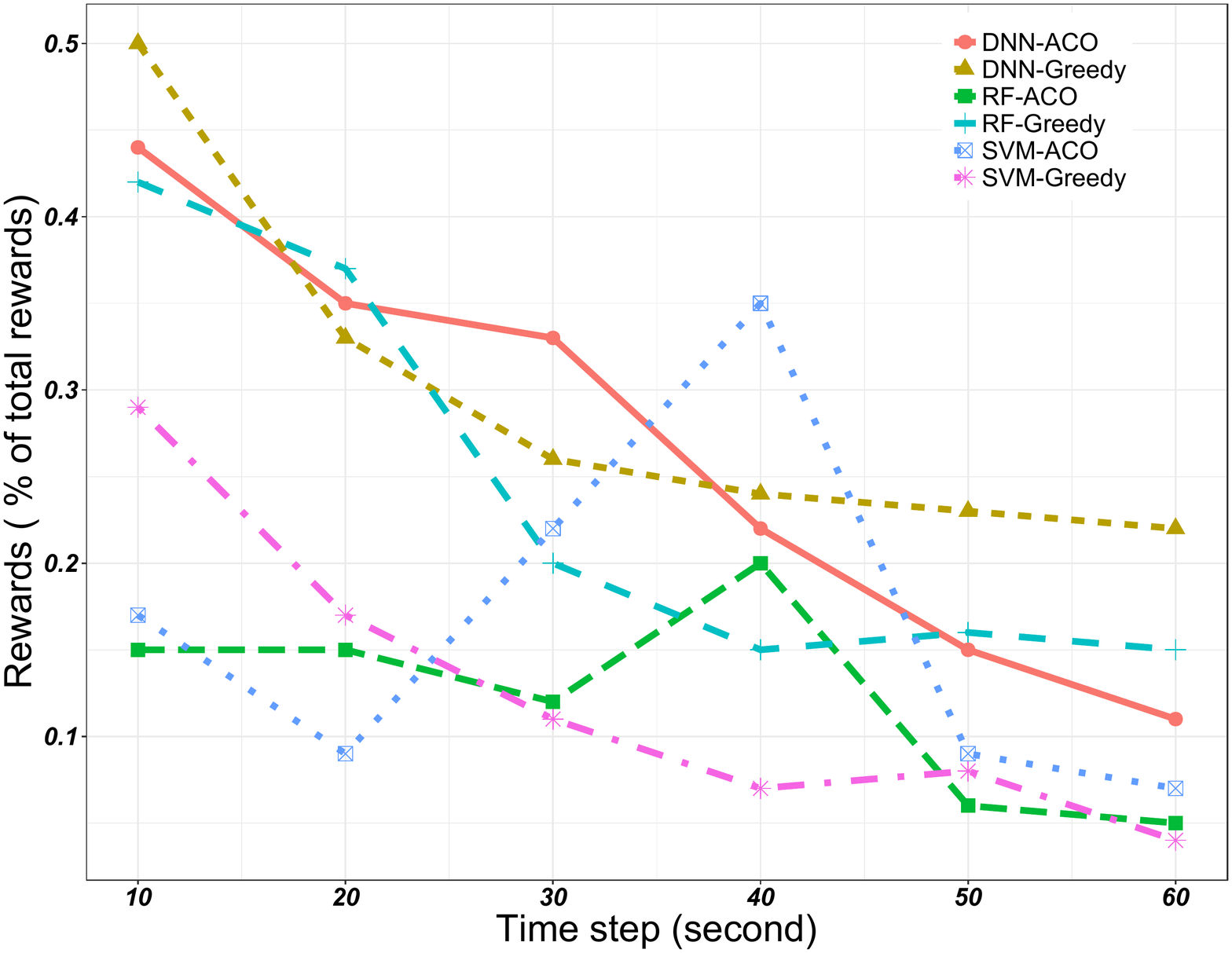}
		\label{fig:rewards_timeslots}
	\end{subfigure}%
	\begin{subfigure}[b]{0.49\columnwidth}
		\includegraphics[width=\columnwidth]{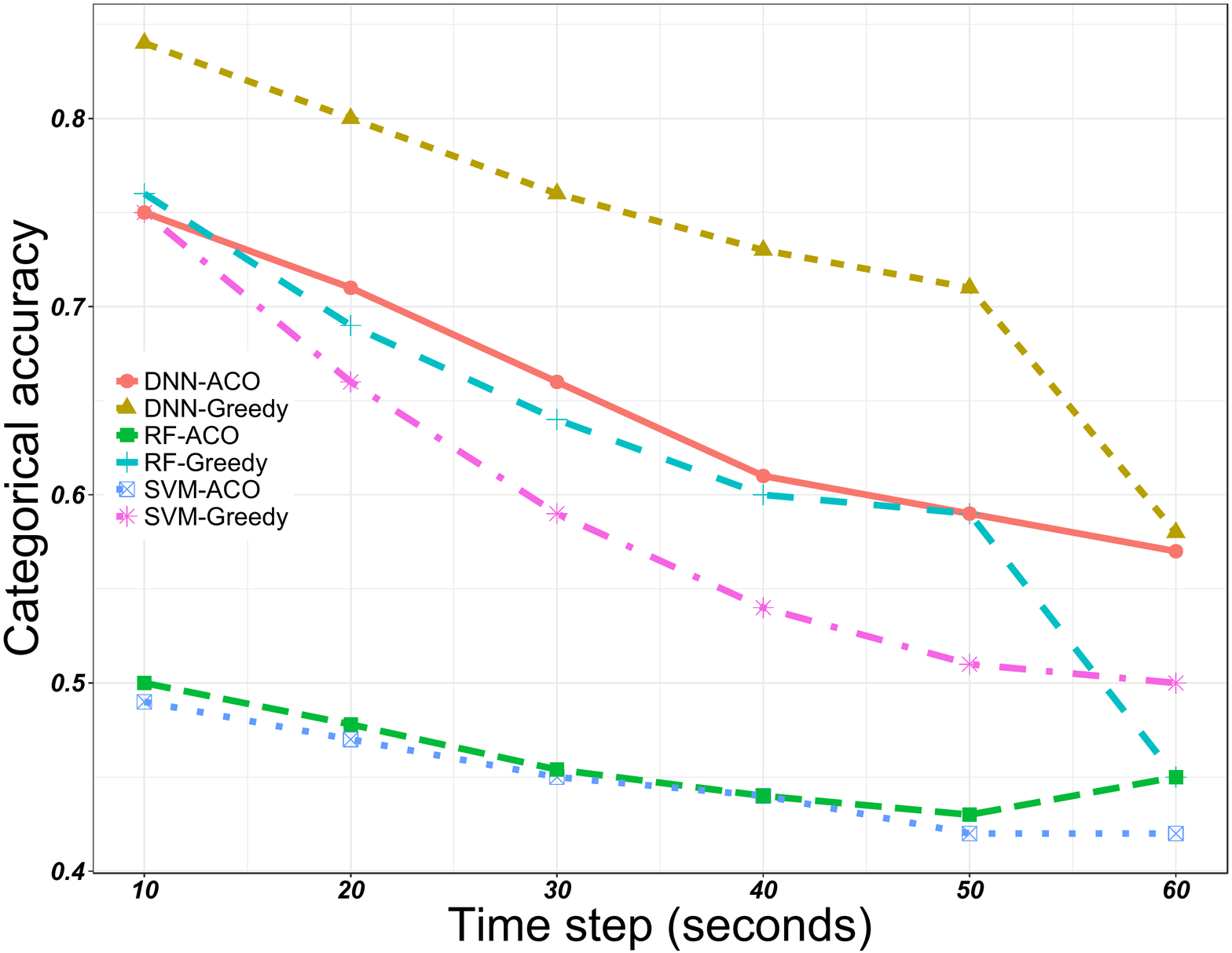}
		\label{fig:acc_timeslots}
	\end{subfigure}%
	\caption{The effect of the time step $\Delta t$ on the total reward obtained, and the approximation accuracy as seen on the test dataset.}
	\label{fig:timeslots}
\end{figure}

We evaluated our classifiers on graphs with sizes varying from $10$ to $50$ nodes. Figure \ref{fig:nodes} shows that the classification accuracy drops significantly as the size of the graph is increased, which suggests that classification methods imitate optimisation solutions well in smaller search spaces. It is possible that in larger search spaces, the limited number of training examples is the limiting factor preventing the classifiers from being able to learn better an approximation of the optimisation routine. Though the rewards increase with the number of nodes in the problem, this is likely the result of greater potential rewards due to the presence of more nodes in the graph. Notably, the gap in performance between the optimisation solution and the classification solutions become larger for these larger problems. The DNN outperformed both the SVM and RF solutions by increasing its total reward along with the graph size. In contrast, the SVM and RF performance dropped under the same conditions. Interestingly, we find that all classifiers achieve higher accuracy when they learn from greedy than ACO. This phenomenon may be caused by the complexity of the optimisation solution.  This is planned for future study. 

\begin{table}[!ht]
\centering
\caption{Runtime for different algorithms with different number of nodes. (seconds)}
\label{tab_time} 
\begin{tabular}{p{0.06\textwidth}|p{0.06\textwidth}p{0.06\textwidth}p{0.06\textwidth}p{0.06\textwidth}}
\hline 
  \#Nodes & Greedy & DNN-Greedy & ACO & DNN-ACO \\
\hline 
 \multirow{4}{*}{\makecell[l]{10\\20\\30\\40\\50}} & \multirow{4}{*}{\makecell[l]{3.12s\\6.99s\\9.23s\\11.99s\\16.10s}} & \multirow{4}{*}{\makecell[l]{0.24s\\0.57s\\1.04s\\1.33s\\1.72s}} & \multirow{4}{*}{\makecell[l]{74.89s\\152.51s\\231.50s\\323.39s\\405.44s}} & \multirow{4}{*}{\makecell[l]{0.27s\\0.45s\\0.89s\\1.21s\\1.53s}} \\
  &   &   &   &   \\
  &   &   &   &   \\
  &   &   &   &   \\
  &   &   &   &   \\
 \hline
\end{tabular}
        \end{table}

We also evaluated the model on data from a single day with varying time step sizes from $10$ to $60$ seconds. Figure \ref{fig:timeslots} shows that smaller time steps resulted in better overall accuracy on the validation set, but this did not necessarily translate to better rewards. This is potentially because smaller time steps provided more training data. However, it does not suggest that testing accuracy is higher.

\subsection{Computational Complexity Analysis}
Finally, we evaluated the computational efficiency of both the optimisation method and the neural network, by varying the number of nodes from $10$ to $50$ and then measuring the execution time of the program. Figure \ref{tab_time} shows that the running time for customised DNN is much faster than the optimisation algorithm since we exclude the training time. We only consider the test session time as the running time of DNN because the training session can be done offline. That is, we can use historical data to train the DNN before using it. Optimisation methods cannot be applied to the real scenario before we know it. Therefore, it is fair to compare the testing time of DNN and running time of the optimisation methods.

It shows that testing time is significantly shorter than other algorithms and does not change with the number of nodes. Therefore, DNN-based model is much more efficient than optimisation solution in the real-world scenario.

\section{Discussion and Future Work}
\label{chapter:DiscussionAndFutureWork}

The neural network performed reasonably consistent across all days, though in general, it fared worse compared to the greedy algorithm during the week. Notably, we found that when the performance of the greedy algorithm was low, the gap between the neural network and the optimisation algorithm was very small.

In this paper, we only design a customised DNN for the greedy algorithm which may not suitable for ACO and other optimisation algorithms such as greedy algorithm is one of the simplest optimisation algorithms. Our next goal is to design different neural networks architectures for different optimisation methods and generalise current DNN to more existing optimisation problems.
\section{Conclusion}
\label{chapter:conclusion}
A technique was shown for reformulating an orienteering problem as a classification problem, and using conventional optimisation to generate labels for training. We took finer time slices of the dataset to increase the amount of training data, and this was shown to improve accuracy. We are the first group who design a neural network to approximate the greedy algorithm in TOP. We evaluated on a large real-world dataset by sampling at a different time interval to generate a distinct test set. It was shown that our customised DNN could be used to approximate the greedy algorithm in TOP.

\bibliographystyle{named}
\bibliography{ijcai19}

\end{document}